\documentclass[submission,copyright,creativecommons]{eptcs}
\providecommand{\event}{IJCAI-AREA 2021} % Name of the event you are submitting to
\usepackage{breakurl}             % Not needed if you use pdflatex only.
\usepackage{underscore}           % Only needed if you use pdflatex.
\usepackage{graphicx}
\usepackage{times}
\usepackage{soul}
\usepackage{url}
\usepackage{graphicx}
\usepackage{amsmath}
\usepackage{amsthm}
\usepackage{booktabs}
\usepackage{iftex}

\usepackage{underscore}         % Only needed if you use pdflatex.
\usepackage[T1]{fontenc}        % Recommended with pdflatex

\usepackage{acronym}
\usepackage{subfig} % for subplots

\usepackage{dirtytalk}

\usepackage{multirow}
\usepackage{tabularx}
\usepackage{changepage}
\usepackage[absolute]{textpos}
\usepackage{siunitx}
\usepackage{multicol}
\usepackage{fancyvrb}
\usepackage{listings}
\usepackage{verbatimbox}
\usepackage{amsmath}
\usepackage{dirtytalk}
\usepackage{xcolor}

\usepackage{amssymb}
\usepackage{amsfonts}
\usepackage{amstext}
\usepackage{amsthm}
\usepackage{adjustbox}

\usepackage[ruled,vlined,linesnumbered]{algorithm2e}
\usepackage{xcolor}
\usepackage{algorithmicx}
\usepackage{algcompatible}
\usepackage[noend]{algpseudocode}
\usepackage{rotating,color}

\SetCommentSty{mycommfont}

\usepackage[utf8]{inputenc}
\usepackage{varwidth}

\usepackage[skins]{tcolorbox}
\usepackage{algorithmicx}
\usepackage{algcompatible}
\usepackage[noend]{algpseudocode}
\usepackage{tabularx}
\usepackage{adjustbox}
\usepackage{listings}
\SetKwInput{KwInput}{Input} % Set the Input
\SetKwInput{KwOutput}{Output} % set the Output

\usepackage{verbatimbox}
\usepackage{graphicx}
\usepackage{subfig}
\usepackage{xcolor}

\usepackage{mathtools}

\usepackage{color,colortbl}

\SetNlSty{textbf}{\color{black}}{}

\SetCommentSty{mycommentfont}

\usepackage{dirtytalk}
\usepackage{tcolorbox}

\definecolor{MyDarkBlue}{rgb}{0.15,0.25,0.45}
\definecolor{applegreen}{rgb}{0.55, 0.71, 0.0}
\definecolor{burntorange}{rgb}{0.8, 0.33, 0.0}
\definecolor{burntumber}{rgb}{0.54, 0.2, 0.14}
\definecolor{almond}{rgb}{0.94, 0.87, 0.8}
\definecolor{antiquewhite}{rgb}{0.98, 0.92, 0.84}
\definecolor{antiquebrass}{rgb}{0.8, 0.58, 0.46}

\newtheorem{example}{Example}

\newtheorem{definition}{Definition}
\newtheorem{lemma}{Lemma}

\title{Temporal Planning with Incomplete Knowledge and Perceptual Information}

\author{Yaniel Carreno\thanks{Contact Author.} \; \;  Yvan Petillot \; \; Ronald P. A. Petrick%$^{1,2,3}$\footnote{Contact Author}\and
\institute{Edinburgh Centre for Robotics, Edinburgh, United Kingdom}
\institute{Heriot-Watt University and The University of Edinburgh, Edinburgh, United Kingdom}
\email{\{y.carreno, y.r.petillot, r.petrick\}@hw.ac.uk}
}
\def\titlerunning{Temporal Planning with Incomplete Knowledge and Perceptual Information}
\def\authorrunning{Y. Carreno, Y. Petillot, R. Petrick}
\begin{document}
\maketitle

\begin{abstract}
%In real-world applications, the ability to reason about incomplete knowledge, sensing, temporal notions, and numeric constraints is vital. While several AI planners are capable of dealing with some of these requirements, they are mostly limited to problems with specific types of constraints. This paper presents a new planning approach that 
%supports temporal constraints and %contingency reasoning. The new solver combines contingent plan construction within a temporal planning framework, offering solutions that consider numeric constraints and incomplete knowledge. We propose a small extension to the Planning Domain Definition Language (PDDL) to model (i) sensing actions that
%reveal the possible value of  operate over unknown propositions, (ii) incomplete knowledge, and (iii) possible outcomes from non-deterministic sensing action. We also introduce a new set of planning domains to evaluate our solver, which has shown good performance on a variety of problems.

In real-world applications, the ability to reason about incomplete knowledge, sensing, temporal notions, and numeric constraints is vital. While several AI planners are capable of dealing with some of these requirements, they are mostly limited to problems with specific types of constraints. This paper presents a new planning approach that 
%supports temporal constraints and %contingency reasoning. The new solver
combines contingent plan construction within a temporal planning framework, offering solutions that consider numeric constraints and incomplete knowledge. We propose a small extension to the Planning Domain Definition Language (PDDL) to model (i) incomplete, (ii) knowledge sensing actions that operate over unknown propositions, and (iii) possible outcomes from non-deterministic sensing effects. We also introduce a new set of planning domains to evaluate our solver, which has shown good performance on a variety of problems.
\end{abstract}

\section{Introduction and Motivation}

Automated planning is widely used as a tool for autonomous agents to achieve their mission goals. In the past decade, AI planners have been introduced in a large number of real-world applications \cite{kunze2018artificial} to deal with the challenges that arise in these scenarios, such as the incompleteness of the domain definition and the complexity of dynamic models that capture numeric and temporal constraints. Hybrid temporal planners are able to reason with both discrete and continuous numeric changes over time to generate
realistic action schedules (plans) capable of supporting
%provide knowledge about the action schedule to generate more realistic plans that support
concurrent actions, synchronisation of multiple tasks, and deadlines. Several solutions to planning with continuous and discrete effects \cite{hoffmann:jair-03,long2003exploiting,coles2010forward} base their reasoning on deterministic models. These solvers generate plans by scheduling sequences of actions that satisfy numeric and temporal constraints. However, their applicability is limited when parts of the domain are incomplete or unknown.  Contingent planning \cite{peot1992conditional,weld1998extending} copes with certain types of incomplete information by treating the plan as a decision tree with different contingent branches that could arise. A contingent plan guides the agent to act conditionally to achieve the goal, with actions in the decision tree enabling the planner to decide which branch to take. 

In this work, we focus on temporal planning with numeric constraints where the action sequences required to reach the goals give rise to conditional plans. For instance, consider the following example:

%, where action sequences meet propositional preconditions. The applicability of such planners is limited when parts of the domain are incomplete or unknown, and the planner must consider a set of possible values for certain state features. %Replanning is often considered as a process for dealing with unexpected changes in the world. However, replanning requires additional computational effort and introduces delays due to the generation of new plans, which is not optimal, particularly when the initial state of the problem is modelled as a set of
%in problems where the modelling of the initial state can include 
%\emph{possible states} of the physical environments or the behaviour of a system. Contingent planning \cite{peot1992conditional,weld1998extending} copes with certain types of incomplete information by treating the plan as a decision tree with different contingent branches that could arise. A contingent plan guides the agent to act conditionally to achieve the goal, with actions in the decision tree enabling the planner to decide which branch to take. In this paper, we focus on temporal planning problems with numeric constraints where the action sequences required to reach the goals give rise to conditional plans. For instance, consider the following example:

\begin{figure}
\centering
\includegraphics[width=0.49\columnwidth]{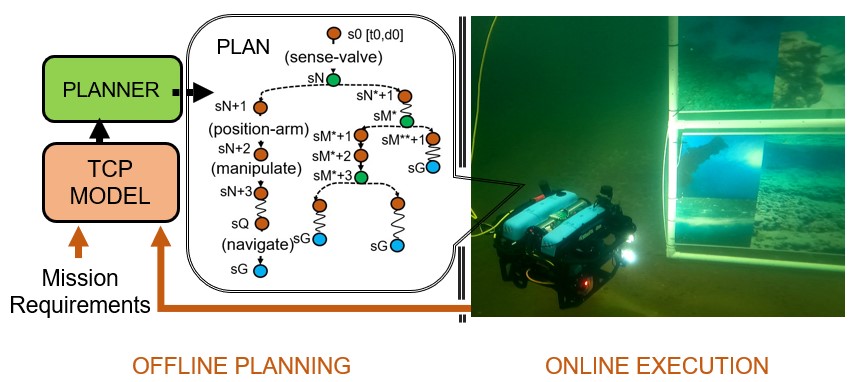}
\includegraphics[width=0.49\columnwidth]{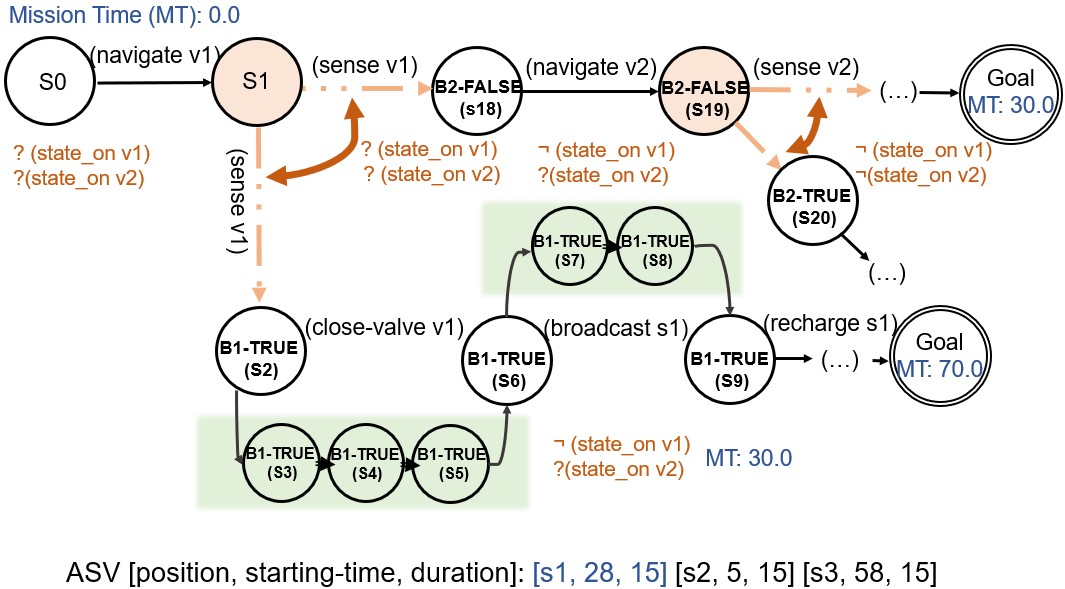}
\caption{\textbf{(left)} Example of the planning and execution process required to solve a problem in Example~\ref{example}.
%Representative example of the planning and execution process to solve the Example \ref{example} problem.
%Offline planning reasons about domain incompleteness, temporal  and numeric constraints to  generate an offline plan solution. During the online execution phase the perception information is used to decide the sequence of actions to implement (branch) based on the knowledge acquired at the runtime.
\textbf{(right)} Graphical representation of the plan solution for Example \ref{example}. 
}
\label{fig: chapter:6 tcp system-framwork-initial}
\end{figure}

\begin{example}
\label{example}
 \textit{(Valve Manipulation)} An offshore scenario (Figure \ref{fig: chapter:6 tcp system-framwork-initial} (left)) includes a set of blowout preventers (BOPs) which are controlled by manipulating valves. An autonomous underwater vehicle (AUV) must close two valves (v1 and v2) during a mission, and must record and communicate data every time it manipulates a valve. The AUV starts at the deployment base. From there, it can navigate to the BOPs and manipulate a valve. The AUV's actions will depend on the valve state: if a valve is open, then it should be closed; if a valve is closed, then no action is needed. The valve state can be checked using a sensing action. The AUV may also need to recharge using an autonomous surface vehicle (ASV) which is deployed in different recharge points at different times.
\end{example}

Figure \ref{fig: chapter:6 tcp system-framwork-initial} (right) shows a plan solution for this problem. We note that no sequence of actions allows the AUV to achieve the goal without first determining the state of the valves. As a result, choosing the correct actions to execute depends on the outcome of sensing the valve states (open or closed). Numeric constraints are required to control the data recorded when a valve is manipulated. Therefore, if the valve is open, the AUV should execute a manipulation action and navigate to the surface to report the action's implementation. Temporal constraints are also essential for scheduling recharge activities due to the ASV's time availability at different locations. Different sensing outcomes (red) can lead to different task sequences with extra actions (green) requiring more energy and time to complete the mission. 

%For instance, the two subplans in Figure~\ref{fig: chapter:6 tcp system-framwork-initial} reach the goal state with a time difference of 40 min. Following the first branch (top), the plan solution assumes that both valves are closed. Therefore, the number of intermediate actions in the mission is reduced. On the other hand, the bottom branch describes the case when both valves are open, and the AUV needs to implement a set of intermediate actions (green) associated with the manipulation and the broadcasting. If the AUV follows the top branch at the runtime based on the sensing actions outcomes, the robot does not need to recharge. However, if the robot takes the bottom branch, the energy consumption forces recharging during the mission. This is not trivial, as the AUV needs to coordinate its efforts to find the right time slot (blue) considering the ASV's location and time availability. In the example, the ASV is at waypoint \textit{s1} and time 30 min (after the mission started). Then the recharging is implemented. The times and location the robot can recharge might change around the branches depending on the AUV's battery consumption, the ASV's location and time availability.

In this paper, we formalise  temporally-contingent planning (TCP) problems, which describe a partially observable environment with sensing actions (PPOS) and time constraints, and introduce
a new compilation-based planning approach called Temporal Contingent Effect (TraCE) that combines temporal plan construction with contingent planning.
TraCE analyses the PPOS as a Fully-Observable Non-Deterministic (FOND) planning problem using a modified version of the approach in \cite{bonet2011planning} to convert PPOS into FOND problems that include temporal constraints.
Both deterministic (physical) actions and non-deterministic (sensing) actions are treated as durative actions. Plan branches are computed by considering the branching points, arising from sensing actions, and branch depth. The plans generated by TraCE are trees of actions (Figure \ref{fig: chapter:6 tcp system-framwork-initial} (right)), where the actions in a branch satisfy the ordinary propositional preconditions but are scheduled by a sub-solver that also considers temporal and numeric requirements. The approach is evaluated on a set of planning problems that include real robotic applications. 

\section{Related Work}

Many temporal planning models introduce an explicit notion of time \cite{ghallab2004automated}. Examples include solvers dealing with temporal coordination \cite{eyerich2009using}, continuous effect requirements \cite{coles2010forward}, and preferences \cite{benton2012temporal}.  Such  approaches are capable of solving a large number of well-established planning domains \cite{long20033rd}, and real-world domains \cite{carreno2020aamas,carreno2022aamas,carreno2021situation}. Our framework aims to solve a new set of problems that require reasoning about incomplete information in addition to the temporal and numeric requirements. In this work, we focus on offline planning strategies. Previous examples of offline contingent planning solvers include MBP \cite{bertoli2006strong}, Contingent-FF \cite{hoffmann2005contingent}, and  POND \cite{bryce2006planning}. Although these planners can solve large problems at different hardness levels, they do not scale well, mostly due to the underlying belief state representation. DNFct \cite{to2011contingent} and CLG \cite{albore2009translation} (offline version)  
outperform previous approaches, finding solutions in a shorter time with better scalability. A different contingent planning approach is PKS \cite{petrick2002knowledge,petrick2004extending} which attempts to model the knowledge state directly. Amongst the group of planners that deal with FOND planning problems, PO-PRP \cite{muise2014computing} extends PRP \cite{muise2012improved} to compute strong cycle plans. ProbPRP \cite{camacho2016fond} is an extension of PO-PRP that computes policies to overcome deadends. HCP-ASP \cite{yalciner2017hybrid} and HCPlan \cite{nouman2021hybrid} demonstrate the applicability of contingent planning in hybrid approaches.  Our work inherits some of the ideas introduced by the prior approaches regarding implementing hybrid solutions and the translation (or compilation) of nondeterministic problems for planning.

Few approaches in the literature combine temporal and contingent planning to solve nondeterministic temporal planning problems. CTP \cite{tsamardinos2003ctp} allows the construction of conditional plans with temporal constraints. Strategies based on Simple Temporal Networks with Uncertainty (STNU) such as \cite{cimatti2014using} are used as temporal scheduling tools where conditions and decisions can be added to STNUs \cite{combi2013algorithm,zavatteri2019conditional}. \cite{combi2019conditional} present an encoding to Conditional Simple Temporal Networks with Uncertainty and Resources (CSTNURs) with promising results in a set of applications that model temporal and numeric constraints. A solution that considers planning and meeting temporal problem constraints is the TCP framework \cite{foss2005generating}, which includes time and contingency notions in the model. This framework differs from the contingent planning problems we are interested in, where branch creation leads to solve the incomplete knowledge %where branch creation addresses the incomplete knowledge. 
Some work associated with temporal plan merging \cite{hashmi2010merging,carreno2020towards} and opportunistic planning consider temporal \cite{cashmore2017opportunistic} and resource \cite{coles2012opportunistic} constraints to generate branches in a temporal plan. \cite{carreno2021compiling} presents an approach to solve TCP problems considering the work implemented in \cite{palacios2014compiling} that translates 
contingent planning into classical problems. TraCE differs from this approach in the problem translation and the tree expansion.

\section{Problem Formalisation}
\label{sec: formalisation}

We begin by formulating the TCP problem. We use
%review the policy solving the model. To do so, we revise first the temporal and contingent planning problems 
PDDL2.1 \cite{fox2003pddl2} which is typically used for temporal planning problems:

\begin{definition}
\label{def: chapter3: tp problem}
A temporal planning problem is a tuple  ${{P}_{T}:= \langle {P}, {V}, {A}_{T}, {I}_{T}, {G}_{T}, {T} \rangle}$, where  ${P}$ is a set of Boolean variables; ${V}$ is a vector of real variables (numeric fluents);  ${A}_{T}$ is a set of instantaneous and durative actions, where the duration of the actions is controllable and known; ${I}_{T}$ is a function
%over ${P} \cup  {V}$ which describes the initial state,
${I}_{T}: {P} \cup {V}  \rightarrow \{ \top, \bot \} \cup \mathbb{R}$ describing the initial state; ${G}_{T}$ is a set of goals, where goals in ${G}_{T}$ are Boolean variables from ${P}$ or inequalities over real variables in ${V}$ that represent the objectives that must be achieved; ${T}$ is a set of timed initial literals (TILs).
\end{definition}

\noindent
%We define ${T}$ as the set of TILs.
A timed initial literal (TIL) \cite{cresswell2003planning} is a  pair (\textit{t}, \textit{p}) where \textit{t} is the rational-valued time of occurrence of a Boolean variable $p$, where $p \in {P}$ and  $\textit{p}  \rightarrow \{p, \neg p\}$. A TIL $(t, p)$ defines the time $t$ that the Boolean variable $p$ becomes true ($p$), and a TIL  $(t, \neg{p})$  describes the time $t$ that $p$ becomes false ($\neg p$).
%$(t_{i}, p_{i})$ defines the time $t_{i}$ that the Boolean variable $p_{i}$ becomes true ($p_{i}$), and a TIL  $(t_{i}, \neg{p_{i}})$  describes the time $t_{i}$ that $p_{i}$ becomes false ($\neg p_{i}$).

\begin{definition}
\label{def: chapter3: durative action}
A durative action $a_d \in {A}_{T}$ is a tuple of the form $\langle a_{d_{\mathit{pre}}}, a_{d_{\mathit{eff}}}, a_{d_{\mathit{dur}}} \rangle$.
$ a_{d_{\mathit{pre}}}$ is a set of conditions of the following types that must hold for the action to be applicable: at-start ($ {a}_{d_{{\mathit{pre}}}\vdash}$), over-all ($ {a}_{d_{{pre\leftrightarrow}}}$), and at-end ($ {a}_{d_{{\mathit{pre}\dashv}}}$). $a_{d_{\mathit{eff}}}$ is the set of action effects of the types: positive starting effects (${a}_{d_{{\mathit{eff}\vdash}^{+}}}$), negative starting effects (${a}_{d_{{\mathit{eff}\vdash}^{-}}}$), numeric starting effects (${a}_{d_{{\mathit{eff}\vdash}^{n}}}$), continuous numeric effects (${a}_{d_{{\mathit{eff}\leftrightarrow}^{n}}}$), positive ending effects (${a}_{d_{{\mathit{eff}\dashv}^{+}}}$), negative ending effects (${a}_{d_{{\mathit{eff}\dashv}^{-}}}$), and numeric ending effects (${a}_{d_{{\mathit{eff}\dashv}^{n}}}$).  $a_{d_{\mathit{dur}}}$ is a set of action durations.
\end{definition}

%The ${\Pi}_{T}$ is called time-aware plan and specifies the solution of a temporal planning problem as follows:

%\begin{definition}
%\label{def: chapter3: time-aware plan}
%A time-aware plan  ${\Pi}_{T}$ for ${P}_{T}$ is a set of tuples ${{\pi}_{T_{i}} := \langle a, t, d\rangle}$, where $a \in {A}$ is the action, $t$ its starting time and $d$ the action's duration. A durative action $a$ hold  $t \in 	{\mathbb{R}}_{\geq0}$ and $d \in {\mathbb{R}}_{>0} $,  where ${\mathbb{R}}_{\geq0} = \{ x \in \mathbb{R} \, | \, x \geq 0\}$ and ${\mathbb{R}}_{>0} = \{ x \in \mathbb{R} \, | \,  x > 0\}$.
%\end{definition}

A solution to a TCP problem
${P}_{T}$ is a time-aware plan
${\Pi}_{T}= \{a_{1} \cdots, a_{n}\}$ described by a sequence of durative and instantaneous actions, where each action $a_i$ is applicable, and $\Pi_{T}$ achieves the goals in ${G}_{T}$ satisfying the temporal constraints. For instance, each branch in Figure~\ref{fig: chapter:6 tcp system-framwork-initial} (right) from the initial state ${I}_{T}$ to the goal ${G}_{T}$ is a single sequence of durative actions. Note that such plans do not consider incomplete information or the effects of sensing actions.

We now consider contingent planning problems which include partial observability and non-deterministic (sensing) actions in the model, but no temporal constraints.

\begin{definition}
\label{def: chapter3: contingent-planning-problem}
A contingent planning problem is a tuple ${{P}_{C} := \langle {P}, {A}_{C}, {O}, {I}_{C}, {G}_{C}\rangle}$, where   ${P}$ is set of Boolean  variables  describing the state of the world; ${A}_{C}$ is a set of physical actions; ${O}$ is a set of %observations 
sensing actions, separate from ${A}_{C}$ such that ${O} \cap {A}_{C} = \emptyset$;  ${I}_{C}\subseteq {P}$, is the set of clauses over ${P}$ that denotes the initial state; and ${G}_{C}$ is a set of propositions over ${P}$ representing the goal condition.
\end{definition}

A proposition $p \in {P}$ or its negation $\neg p$ defines a \textit{literal} $l \in {L}$, where ${L}$ is a set of literals and $\overline{{L}}= \{ \neg l \;| \;l \in {L}\}$. A set ${L}^{'}$ of literals is  consistent if the condition $\{p, \neg p \} \nsubseteq {L}$ holds for every $p$; and complete if the condition $\{p, \neg p\} \cap {L} \neq \emptyset$ holds for every $p \in {P}$. The complement of a literal $l$ is defined as $\overline{l}$ such that $\overline{l}= \neg l$ and $p=\neg \neg p$ for $p \in {P}$.

A \textit{state} $s$ is defined as a consistent and complete  set ${L}$ of literals. A state $s$ satisfies a conjunction of literals ${L''}$ ($s \models {L''})$ if ${L''} \subseteq s$. A state $s$ satisfies a literal $l$ $(s \models l)$ if $l \in s$. A \textit{belief state} ${S}$ is a set of states. The belief state ${S} \models l$ if $s \models l$ for every $s \in {S}$. Finally, ${S} \models {L''}$ if $s \models {L''}$ for every $s \in {S}$.

%In a contingent planning model, the states in the belief state ${S}$ capture the possible outcomes of an \say{unknown} literal. The set of states in the belief state capture the alternative ways that the world could be configured.

The set of states in the belief state ${S}$ capture the alternative ways that the world could be configured. In a contingent planning model, a literal as being \say{unknown} if it appears in more than one state in a belief state  with different mappings. So, if $s \models l$ is true in each state $s$, then $s$ is said to be known. However, if there exist states $s_{1}$ and $s_{2}$ such that $s_{1} \models l$ and $s_{2} \models \neg l$ then $l$ would be considered unknown since either value of $l$ is considered possible. For instance, in Example \ref{example} a belief state ${S}$ includes the possible outcomes of the {\footnotesize\texttt{(state\_on v1)}} variable which are  {\footnotesize\texttt{(state\_on v1)}} and $\neg${\footnotesize\texttt{(state\_on v1)}}. 

Actions are defined as follows:

\begin{definition}
\label{def: chapter3: action}
A physical action $a \in {A}_{C}$ is a tuple  $\langle a_\mathit{pre}, a_\mathit{eff}\rangle$, where $a_\mathit{pre}$ is a set of atomic propositions indicating the preconditions to implement $a$, and $a_\mathit{eff}$ represents a set of action outcomes. $a_\mathit{eff}$  describes  a set of pairs $\langle c,l\rangle$,% which capture its conditional effects, where $c$ is a set of conditional effects defined by a set of literals ${L}$ and $l$ is a literal.
which capture its conditional effects, where $c$ is a set of possible effects defined by a set of literals ${L}$ and $l$ is a literal.
\end{definition}

\begin{definition}
\label{def: chapter3: observations}
 %An observation 
 A sensing action $o \in {O}$ is a tuple  $\langle {o}_\mathit{pre}, {o}_\mathit{eff} \rangle$, where  ${o}_\mathit{pre}$ is a set of atomic propositions indicating the preconditions to implement $o$, and ${o}_\mathit{eff}$ completely characterises an  unknown literal $l$, % in the initial state ${I}_{{C}}$, where $l \in {I}_{{C}}$,  by considering the possible outcomes for $l$ described in a belief state ${S}$. Therefore, ${o}_\mathit{eff}$ includes a set of possible effects ${o}_\mathit{eff}'$ and ${o}_\mathit{eff}''$ that uncover the truth value of  $l$. This is for a belief state ${S}$ a sensing action effects define two independent belief state ${S}^{+}$ and ${S}^{-}$ that enclose the states where $l$ is possible, and states where $\neg l$ is possible, respectively.
 by considering the possible outcomes for a given proposition. Therefore, ${o}_\mathit{eff}$ includes a set of possible effects ${o}_\mathit{eff}'$ and ${o}_\mathit{eff}''$ that uncover the value of $l$.
\end{definition}

This is for a belief state ${S}$ a sensing action effects define two independent belief state ${S}^{+}$ and ${S}^{-}$ that enclose the states where $l$ is possible, and states where $\neg l$ is possible, respectively. Considering Definition \ref{def: chapter3: observations}, sensing actions are nondeterministic. Sensing actions lead to a possible set of outcomes (states), enclosed in the belief state, associated with the truth value of a literal. For a sensing action the belief state ${S}_{o}$ gives rise to two possible belief states, depending on the value of the underlying literal ($l$ or $\neg l$). %underlying literal in the real world ($l$ or $\neg l$). 
This makes the model nondeterministic at plan time but the belief state is always certain in each case. The possible belief states associated with a sensing action are: ${S}^{+}_{o} = \{ s \, | \, s \in {S}, s \models l(o)\}$ and ${S}^{-}_{o} = \{ s \, | \, s \in {S}, s \models \overline{l(o)}\}$.

A solution to ${\Pi}_{C}$ is a 
contingent-aware plan ${P}_{C}$ with a branching structure induced by the sensing outcomes. For instance, the plan in Figure~\ref{fig: chapter:6 tcp system-framwork-initial} (right) is a solution to a contingent planning problem, assuming the recharge and broadcast actions, which include temporal and numeric constraints, are ignored. 

\subsection{Temporally-Contingent Planning Problem}

A TCP problem is a special case of the PPOS problem where the %with uncertainty the
source of uncertainty is connected to the outcome of the sensing actions \cite{muise2014computing}, and temporal and numeric notions are introduced in the model. We consider simple PPOS problems \cite{bonet2011planning} which can be mapped into FOND planning problems \cite{muise2014computing}. In particular, we assume that (i)  non-unary clauses in the initial state ${I}$ are invariant, which are states that hold in every world state \cite{helmert2009concise},  (ii)  hidden literals do not emerge in the effects of a non-deterministic action, and (iii) uncertainty decreases monotonically, i.e., unknown properties cannot become unknown again after becoming known.
%These three properties encloses the uncertainty associated with the nondeterministic action outcomes in TCP problems.
As a result, the TCP problem is a FOND problem with temporal requirements and real variables.  

\begin{definition}
\label{def: TCP}
A temporally-contingent planning problem is a tuple ${P}_{TC} := \langle {P}, {V}, {A}, \Delta, {I}, {G}, {T} \rangle$, where   ${P}$ is set of Boolean  variables  describing the state of the world; ${V}$ is a vector of real variables (numeric fluents), ${A}$ is a set of  durative physical (deterministic) actions; $\Delta$ is a set of  durative sensing (nondeterministic) actions, separate from ${A}$ such that $\Delta \cap {A} = \emptyset$,  both actions with controllable and known durations; ${I}$ is a function
%over ${P} \cup  {V}$ which describes the initial state,
${I}: {P} \cup {V}  \rightarrow \{ \top, \bot \} \cup \mathbb{R}$ describing the initial state; ${G}$ is a set of goals, where goals in ${G}$ are Boolean variables from ${P}$ or inequalities over real variables in ${V}$ that represent the objectives that must be achieved; 
${T}$ is a set of TILs.
\end{definition}

The TCP problem includes two type of actions, where physical actions can be defined as  durative actions (see Definition \ref{def: chapter3: tp problem}),  while durative sensing actions in $\Delta$ are an special type of sensing actions (see Definition \ref{def: chapter3: contingent-planning-problem}) defined as follows:

\begin{definition}
\label{def: sensing-action}
A durative sensing action $\delta$, where $\delta \in \Delta$, is a durative action with a nondeterministic outcome defined  by the tuple $\langle \delta_\mathit{pre}, \delta_\mathit{eff}, \delta_\mathit{dur}\rangle $. $\delta_\mathit{pre}$ is a set of conditions, including a set of literals, that must hold for the action to be applicable of type: at-start, over-all, and at-end. $\delta_\mathit{eff}$ includes a set of possible action outcomes $\delta_\mathit{eff}'$ and $\delta_\mathit{eff}''$ that include  effects of positive, negative, and numeric startings, continuous numeric, positive, negative, and numeric endings. The possible action outcomes  characterise an  unknown literal $l$, by considering the possible outcomes  for $l$ ($l$ or $\neg l$) described in a belief state ${S}$.
\end{definition}

 In a belief state ${S}_{A} = \{ (l, l_{1}, l_{2}), (\neg l, l_{1}, l_{2}), (l, \neg l_{1}, l_{2}), (\neg l, \neg l_{1}, l_{2}) \}$, there would be states where $l$ is possible and states where $\neg l$ is possible so $l$'s true value is unknown. % Considering $\delta$'s outcomes $\delta_\mathit{eff}'$ and $\delta_\mathit{eff}''$ define a partitioned belief state ${S}_{B} = \{ (l, l_{1}, l_{2}), (l, \neg l_{1}, l_{2}) \; | \; (\neg l,$ $ l_{1}, l_{2}), (\neg l, $ $\neg l_{1}, l_{2}) \}$
 $\delta$'s outcomes $\delta_\mathit{eff}'$ and $\delta_\mathit{eff}''$ define two independent belief states  ${S}_{B}' = \{ (l, l_{1}, l_{2}), (l, \neg l_{1},$ $ l_{2})\}$ and  ${S}_{B}'' = \{ (\neg l, l_{1}, l_{2}), (\neg l, $ $\neg l_{1}, l_{2}) \}$. % The sensing action picks out all the states that are consistent with the sensed literal.
 $\delta_\mathit{dur}$ represents represents a set of duration constraints (controllable and known). A sensing action $\delta$, where $\delta \in \Delta$, helps to completely characterise an  unknown literal $l$, where $l \in {I}$, by considering the possible outcomes for $l$ described in a belief state ${S}'$. The solution of the TCP problem is  a transition tree which is defined based on the outcomes of the sensing action.

\medskip\noindent
\textbf{PDDL Encoding.} We propose a set of extensions to PDDL2.1, which define the unknown literals, the set of possible outcomes of a durative sensing action, and the connection between the literal sensed and the effects of the sensing action. This approach is inspired by previous approaches that represent incomplete information, and sensing  \cite{hoffmann2005contingent,petrick2002knowledge}. Our PDDL extensions to encode the TCP problem are: (i)  {\small\textbf{:unknown-literals.}} Figure~\ref{fig: sensing-action} (left top) shows an example of how incomplete information is represented for Example \ref{example}. In {\footnotesize\texttt{:unknown-literals}}, the incomplete information in the problem is identified, which can be extracted for the initial state ${I}$. For instance, the domain the unknown literals reflect that the valve states are initially unknown, {\footnotesize\texttt{(state\_on ?v - valve)}}; (ii) {\small\textbf{:knowledge-updates.}} Figure~\ref{fig: sensing-action} (left bottom) shows an example of the observational effects extension for the Example \ref{example}. In {\footnotesize\texttt{:knowledge-updates}}, the {\footnotesize\texttt{and}} and {\footnotesize\texttt{oneof}} clauses \cite{geffner2013concise} are used to define the set of facts corresponding to the possible outcomes of a sensing action. For the Example \ref{example} domain, the unknown literal  {\footnotesize\texttt{(state\_on v1)}} is associated with two possible effects, first, the valve's state  is open ({\footnotesize\texttt{(state\_on v1)}}), the second, the valve's state is closed ({\footnotesize\texttt{{(not (state\_on v1))}}}).  Notice that in the second effect, the {\footnotesize\texttt{and}} clause is used to specify the state of two literals, {\footnotesize\texttt{(not (state\_on v1))}} and {\footnotesize\texttt{(valve\_closed wp32)}}; and  (iii) {\small\textbf{:durative-action.}} Finally, we present the extension for sensing outcomes  in Figure \ref{fig: sensing-action} (right). In {\footnotesize\texttt{:durative-action}}, a durative sensing action is extended from an ordinary (physical) action to include the {\footnotesize\texttt{:observe}} construct, which connects the action to the sensed literal. In the example, the sensing action ({\footnotesize\texttt{{sense-valve}}}) observes the valve's state ({\footnotesize\texttt{state\_on ?v - valve}}) for a particular valve {\footnotesize\texttt{?v}} which gives rise to multiple possible outcomes, such as those in  Figure~\ref{fig: chapter:6 tcp system-framwork-initial} (right). This action includes time constraints and is not limited to Boolean effects on literals. A durative sensing action in a TCP problem is nondeterministic  based on the sensing outcomes from the {\footnotesize\texttt{:observe}}. However, a sensing action contains another set of deterministic effects that are associated with the {\footnotesize\texttt{:effect}}. For instance, the positive ending effect {\footnotesize\texttt{(at end   (available ?r))}}, which defines the robot is available when the action finishes. Considering this, the set of belief states associated with the outcome of a sensing action contains an effect associated with the unknown literal (different for each outcome) and a set of effects  that are the same for all possible belief states. In Example \ref{example}, the possible outcomes from implementing the sensing action in valve {\footnotesize\texttt{v1}} are: (i) {\footnotesize\texttt{(and \textcolor{burntumber}{(state\_on v1)} (available auv1) (v\_at v1 wp32) ...)}}, and (ii) {\small {\footnotesize\texttt{(and \textcolor{burntumber}{(not (state\_on v1) (valve\_closed wp32)} (available auv1) (v\_at v1 wp32) ...)}}}.

\begin{figure}[t]
\begin{center}
\tcbset{width=(\linewidth-2mm)/2,before=,after=\hfill,
colframe=black,colback=white, arc=0.1mm}
\begin{tcolorbox}[height=4.8cm,boxrule=0.1mm]
\begin{Verbatim}[fontsize=\scriptsize,commandchars=\\\{\}]
\textbf{(:unknown-literals}
 (state\_on v1)  
 (state\_on v2)
\textbf{)}

\textbf{(:knowledge-updates}
 (oneof (state\_on v1) 
        (and (not (state\_on v1))  
             (valve_closed wp32)))
 (oneof (state\_on v2) 
        (and (not (state\_on v2))  
             (valve_closed wp42)))
\textbf{)}
\end{Verbatim}
\end{tcolorbox}
\begin{tcolorbox}[height=4.8cm,boxrule=0.1mm]
\begin{Verbatim}[fontsize=\scriptsize,commandchars=\\\{\}]
\textbf{(:durative-action sense-valve}
 :parameters (?r - robot ?s - sensor 
              ?v - poi   ?wp - waypoint)
 :duration ( = ?duration 5)
 :condition (and 
            (over all (can_sense ?r ?s))
            (...))
 :effect (and 
         (at end (available ?r)) 
         (...))
 \textbf{:observe (and (at end (state_on ?v)))}
\textbf{)}
\end{Verbatim}
\end{tcolorbox}
\end{center}
\caption{PDDL  extensions  for  representing  incomplete  information and observational effects in durative actions using constructs {\footnotesize\texttt{:unknown-literals}} (left top) and {\footnotesize\texttt{:knowledge-updates}} (left bottom). 
}
\label{fig: sensing-action}
\end{figure}

\section{TraCE Planning}

The TraCE planner takes as input a simple PPOS problem with temporal and numeric constraints ${P}_{TC}$ and generates a time-knowledge-aware plan ${\Pi}_{TC}$ defined as follows: 

\begin{definition}
\label{def: plan}
A time-knowledge-aware plan ${{\Pi}_{TC}} = ({N},{E} )$ for a temporally-contingent planning problem ${P}_{TC}$ is a transition tree ${B}$, represented as an AND/OR graph, where nodes ${N}$ are labelled with actions built on a set of tuples, ${{\pi}_{A} := \langle a, t, d\rangle}$ for physical actions and ${{\pi}_{\Delta} := \langle \delta, t, d\rangle}$ for sensing actions; and edges ${E}$ represent the action outcomes, denoting the set of propositions whose value are known after an action execution, where $a \in {A}$ is an instantaneous or durative action,  $\delta \in \Delta$ is a durative sensing action, $t$ is the action starting time,  $d$ represents the action duration, $t \in {\mathbb{R}}_{\geq 0} $, and $d \in 	{\mathbb{R}}_{>0} $ when actions have a duration.
\end{definition}

The plan ${{\Pi}_{TC}}$ arising from Definition \ref{def: plan} describes a tree where physical and sensing actions are encoded as two different types of vertices: (i) $v_{{A}}$, denoting vertices that describe physical actions, where $v(a)$ is a vertex, $v(a) \in v_{{A}}$, and $a$ is a physical action, $a \in {A}$; and (ii)  $v_{\Delta}$, which describes sensing action vertices, where $v(\delta)$ is a vertex, $v(\delta) \in v_{\Delta}$, and $\delta$ is a sensing action, $\delta \in \Delta$. The ${B}$ tree edges capture the action ordering, where an edge $z(v(x),v(y)$) defines that the action denoted by the vertex $v(x)$ is executed before the action denoted by $v(y)$. Physical actions are deterministic; therefore a vertex $v(a)$ has at most a single edge. Sensing actions are nondeterministic; therefore, a vertex  $v(\delta)$, associated with the sensing action $\delta$, is characterised by at least two possible outgoing edges. In this case, the number of edges depends on the possible $\delta$ outcomes.

Algorithm \ref{alg: TRACE}  shows the TraCE planning approach, which returns a time-knowledge-aware plan ${{\Pi}_{TC}}$. TraCE is a PDDL planner that aims to support the PDDL extensions modelling the TCP problem in full. The required inputs to the TraCE solver are the domain ${D}_{TC}$, including physical and sensing actions; the TCP problem ${P}_{TC}$, which defines the initial state; and ${U}$, which includes the incomplete information and observational effects. ${U}$ incorporates the incomplete information in the {\footnotesize\texttt{:unknown-literals}}'s body and the observational effects in the {\footnotesize\texttt{:knowledge-updates}}'s body; while the sensing actions in ${D}_{TC}$ include the extension for sensing outcomes. The branches in  ${{\Pi}_{TC}}$ are time-aware plan solutions ($\Pi_{T}$) that solve a temporal planning problem ${P}_{T}$ under different sets of contingent outcomes arising from domain incompleteness and durative sensing actions. For instance, the solution for Example \ref{example} shows the branches of a ${{\Pi}_{TC}}$. Top and bottom branches represent two different time-aware plans considering the sensing action {\footnotesize\texttt{sense-valve}} outcomes define the valves ({\footnotesize\texttt{v1}} and {\footnotesize\texttt{v2}}) are (both) closed and open, respectively. 

\begin{algorithm}[t]
\DontPrintSemicolon
\footnotesize
\KwOutput{$\Pi_{TC}$ (Branched Temporal Plan)}
\Begin{
 $\Pi_{T} \leftarrow \emptyset$\\
 $\Pi^{*}_{T}, {H} \leftarrow$ \textsc{FindTemporalPlan}(${D}_{TC}, {P}_{TC}, {U}, t_{\mathit{initial}}$)\\
 \If{\text{CheckExistence}($\Pi_{T}^{*}$)}
 {
 
 \While{$\Pi_{T} \neq \Pi_{T}^{*}$}
 {
 $\Pi_{T}, {H} \leftarrow$ Update$(\Pi_{T}^{*}, {H})$\\
 $b, {Q} \leftarrow$ \textsc{BuildBranch}(${D}_{TC}, {U}, \Pi_{T}, {H}$)\\
 $root \leftarrow$ \textsc{ExpandTree}(${D}_{TC}, {P}_{TC}, b, {H}, {U}, {Q}$)\\
 \textbf{return} $\Pi_{TC} \leftarrow root$
 }
 }
 \Else
 {
 \textbf{return} \textsc{Failure}
 }
 }
\caption{\textsc{TraCE Planner} (${D}_{TC}$, ${P}_{TC}$, ${U}$)}
\label{alg: TRACE}
\end{algorithm}

The strategy starts by initialising an empty temporal plan solution  ${{\Pi}_{T}}$ (line 2), which preserves the temporal plan associated with a particular branch. Then, the approach finds an ordinary (initial) deterministic temporal plan $\Pi_{T}^{*}$ (line 3) for the problem that represents the largest temporal plan solution. In this work, the \textit{plan size} is defined as the total number of actions in a plan (that describe each branch). Therefore, the largest plan is the solution with more actions. For instance, in Example \ref{example},  the largest deterministic temporal plan is represented by the bottom branch. \textsc{FindTemporalPlan} computes the largest plan and returns it in $\Pi_{T}^{*}$. In our approach, the initial temporal plan $\Pi_{T}^{*}$  can be updated in parallel to the ${{\Pi}_{TC}}$ search (line 4-9).

The history set ${H}$ for the temporal plan denotes the state $s[i]$ at which $\Pi_{T}[i]$ or $\Pi_{T}^{*}[i]$ is executed, and the time $t[i]$ at which $i$ starts and concludes, where ${H}[i]=\langle s[i], t[i] \rangle$ and $\Pi_{T}[i]$ or $\Pi_{T}^{*}[i]$ defines the \textit{i}th plan's action in $\Pi_{T}$ or $\Pi_{T}^{*}$. ${H}[i+1]$ describes the possible states reached after an execution of $\Pi_{T}[i]$ or $\Pi_{T}^{*}[i]$. Then,  if  $\Pi_{T}^{*}$ exists (line 4), TraCE checks  if $\Pi_{T} \neq \Pi_{T}^{*}$ (line 5). Notice that the condition $\Pi_{T} \neq \Pi_{T}^{*}$ allows the strategy to update the initial temporally-contingent plan solution if after generating a plan ${{\Pi}_{TC}}$, $\Pi_{T}^{*}$ differs from the first $\Pi_{T}$ as a result $\Pi_{T}^{*}$ was optimised. If the strategy finds discrepancies between $\Pi_{T}$ and $\Pi_{T}^{*}$, the first is updated  with the plan in $\Pi_{T}^{*}$ (line 6). Then, the strategy builds a branch $b$ using the \textsc{BuildBranch} method (see Algorithm \ref{alg: BuildBranch}) and expands the tree for a set of \say{tasks} describing the sensing actions in $b$ to obtain \textit{root} (line 7-8) using the \textsc{ExpandTree} method (see Algorithm \ref{alg: tree}). The variable \textit{root} (identified by first action in the plan) describes ${{\Pi}_{TC}}$, and ${Q}$ defines a queue of tasks $\langle n, u \rangle$, where $n \in {N}$ is a sensing node (representing a sensing action) and $u$ is a possible outcome of the sensing action, where $u \in {U}$. Each $q \in {Q}$ introduces different subsets of initial conditions based on the sensing action outcomes  that leads to a new  temporal planning problem ${{P}_{T}}$. After the tree expansion, the  ${{\Pi}_{TC}}$ is returned (line 9). If $\Pi_{T}^{*}$ is not found the planner fails to find a solution for ${P}_{TC}$ (line 10-11).

The \textsc{FindTemporalPlan} method computes temporal plans using the OPTIC \cite{benton2012temporal} planner. TraCE  branch generation considers temporal–numeric planning within a forward state space search framework. \textsc{FindTemporalPlan} does not consider changes in the temporal planning search regarding the OPTIC approach. Instead, the strategy provides the right conditional and time constraint inputs to the propositional planning techniques and the sub-solver used to schedule the action sequence to generate the required time-aware plan. The algorithm finds the plan with the largest size by iterating over the initial state, considering the combinations of incomplete information and observational effects.  We solve the plan length check by including into the problem initial state the set of observational effects that defines the unknown literals in ${U}$ to be false (${S}^{-}_{o}$). Considering this information in the initial state we obtain a plan solution $\Pi_{T}^{-}$. Then, the plan $\Pi_{T}^{-}$ is stored (if it exists). A similar procedure is repeated considering all sensing outcomes associated with the unknown literals are true (${S}^{+}_{o}$) and we achieve a plan solution  $\Pi_{T}^{+}$ for this initial state. The approach compares the two plan solutions ($\Pi_{T}^{-}$ and $\Pi_{T}^{+}$) and its saves the largest plan in $\Pi_{T}^{*}$. The planning problems that we consider include temporal and numeric constraints that force actions in some order in the plan solution. This justifies our decision to start the branch generation using the largest plan as this solution satisfies the problem with the most considerable set of temporal and numeric restrictions. Temporal planners attempt to improve the initial plan solution, usually considering as a metric the makespan minimisation. 

The \textsc{FindTemporalPlan} method returns the plan's history ${H}$. The second component of the ${H}$ is the time $t[i]$ associated with the action execution. The core of the temporal planning solver implements the scheduling of the plan's actions considering STNs and LP methods. We refer to these strategies as a temporal solver (TS). The TS represents the temporal constraints between the set of  time points where actions start or end. Therefore, we can describe the occurrence of an action $i$ in the TS reasoning as the time point when the action starts $i_{\vdash}$ and $i_{\dashv}$ when the action ends. This information is saved in $t[i] = [i_{\vdash}, i_{\dashv}]$. The \textsc{FindTemporalPlan} strategy extends the scheduling analysis by controlling  the $t_{\mathit{initial}}$ value. The $t_{\mathit{initial}}$ is introduced as an input to the approach (see line 3 in Algorithm \ref{alg: TRACE}). For the first temporal plan $\Pi_{T}^{*}$, $t_{\mathit{initial}}$ is considered 0. However, this modification gains importance when expanding the tree based on the sensing vertices. In these cases, the time the sensing action concludes is defined as the $t_{\mathit{initial}}$ for a new branch in the tree associated with a different outgoing edge from the sensing vertex. We put this into context when describing the remaining parts of the algorithm. The TS encodes a TIL by adding a time point $t(p)$ for the occurrence of TIL $p$, with the temporal constraint $t(p) - t_{\mathit{initial}} = time_{TIL}(p)$, where $time_{TIL}(p)$ is the time at which $p$ occurs. The TILs describe other time points that need to be ordered by the TS approach when scheduling the plan's actions.

The \textsc{BuildBranch} method in Algorithm \ref{alg: BuildBranch} describes the branch generation, which starts with the first action in $\Pi_{T}$. The strategy initialises ${Q}$ (line 2) and creates the first branch over the $\Pi_{T}$'s size (line 3). The branch creation relies upon the definition of all nodes in $\Pi_{T}$ (vertices) and their connections (edges). For each action $\Pi_{T}[i]$ in the plan the algorithm initialises a node $n$ (line 4) that includes a set of \textit{definition parameters}: (i) \textbf{$n$.name}. is the name of the action describing the node; (ii) \textbf{$n$.s-action}. Boolean variable to define if the node represents a sensing action; (iii) \textbf{$n$.parent}. is the node $n$ parent; and \textbf{$n$.depth}. number of edges from the root node to $n$.name. The strategy takes the information that the \textsc{FindTemporalPlan} method saves in the history ${H}$ (state and time) to define another two node's parameters (line 5): (iv) \textbf{$n$.state}. defines the state at which the action in $n$.name is executed; and (v) \textbf{$n$.time}. defines the time at which the node (action) starts and concludes. If the plan size is 0 (line 6-7), the branch $b$ contains precisely the node $n$, which contains the set of parameters initialised. On the contrary, the TraCE  creates an outgoing edge from node $n$ to another new node $\textit{child}$ that denotes the next action (line 8-9). Then, the method defines the state and time\footnote{At this stage $t[i+1]$ for an action $i$ includes initial time of the next action which is $i_{\vdash}$ + $\epsilon$.} parameters for the child (line 10). TraCE appends the node $n$ children in  the node parameter (line 11). \textbf{$n$.children} defines the set of $n$.name(s)  (actions) associated with a $n$.name. If the action to label is a sensing action (line 12), TraCE maps the edge from node $n$ to child \textit{child} with the relevant outcome of the sensing action, obtained from the history ${H}$ of $\Pi_{T}$ and ${U}$ (line 13-14). This labelling is stored in the last node $n$ definition parameter:  \textbf{$n$.edge-map}. is the label of the outgoing edge from the node to its child, where $n$ is a sensing node.

The node $n$, describing a sensing action in $\Delta$, and the possible outcomes defined in ${U}$ for the action that does not label any outgoing edge from $n$, describes a new temporally-contingent planning task with an initial state ${I}^{i}$ obtained from ${H}[i]$. The planner stores the sensing nodes and their possible outcomes ($\langle n, u \rangle$) obtained when building the branch $b$ in ${Q}$ (line 15-16) The algorithms returns $b$ and an updated ${Q}$ (line 17). The $n$.depth definition parameter helps to organise the tasks in the queue ${Q}$ to implement the expansion.  

\begin{flushleft}
\begin{minipage}[t]{0.47\textwidth}
\begin{algorithm}[H]
\DontPrintSemicolon
\footnotesize
\KwInput{$\langle {D}_{TC}, {U}, \Pi_{T}, {H} \rangle$}
\KwOutput{$\langle b, {Q} \rangle$}
\Begin{
${Q} \leftarrow \emptyset $\\
\For{$i = 0, \cdots, size(\Pi_{T})-1$ }
{
$n \leftarrow$CreateNode($\Pi_{T}[i]$)\\
$n$.state, $n$.time = ${H}[i]$\\
\If{$i==0$}
{
$b == n$\\
}
\If{$i \neq size(\Pi_{T})-1$}
{
$\textit{child} \leftarrow$CreateNode($\Pi_{T}[i+1]$)\\
$\textit{child}$.state, $\textit{child}$.time  = ${H}[i+1]$\\
$n.$children $\leftarrow$AppendChildToNode$(\textit{child})$\\
}
\If{$n$.\text{s-action}}
{
${U}_{\Delta} \leftarrow$ObtainCurrentOutcome(${H}[i] \cap {U}_{\Pi_{T}[i]}$)\\
$n$.edge-map($\textit{child}$) = $u$\\
\For{$u = {U}_{\Delta}\diagdown\{u_{\Delta}\}$}
{
${Q} \leftarrow$ AddToQueue(${Q}, \langle n, u\rangle)$\\
}
}

}
\textbf{return} $b, {Q}$
}
\caption{\textsc{BuildBranch}}
\label{alg: BuildBranch}
\end{algorithm}
\end{minipage}
\hfill
%\begin{flushright}
\begin{minipage}[t]{0.47\textwidth}
\begin{algorithm}[H]
\DontPrintSemicolon
\footnotesize
\KwInput{$\langle {D}_{TC}, {P}_{TC}, b, {H}, {U}, {Q} \rangle$}
\KwOutput{\textit{root}}
\Begin{
\While {\textit{not}  ${Q}\leftarrow$ $\emptyset$}
{
\For{$\langle n_{i}, u_{i}\rangle \leftarrow$ ExtractFromQueue$({Q})$}
{
${I} = n_{i}.$state\\
${I^{''}} \leftarrow$ ModifyInitialState(${I}, u_{i}$)\\
${P}_{TC}^{''} \leftarrow$ \texttt{update}.${P}_{TC}$(${I^{''}}$)\\
 $\Pi_{T}, {H} \leftarrow$ \textsc{FindTemporalPlan}(${D}_{TC}, {P}_{TC}^{''}, {U}, n_{i}$.time)\\
 \If{\text{CheckExistence}($\Pi_{T}$)}
 {
 $\textit{branch\_child}, {Q}^{''} \leftarrow$ \textsc{BuildBranch}(${D}_{TC}, {U}, \Pi_{T}, {H}$)\\
  $\textit{branch\_child}.$edge-map $= u_{i}$\\
  $n_{i}.$children$\leftarrow$ AddToTree($\textit{branch\_child}$)\\
  ${Q} \leftarrow$\texttt{update}.${Q}$(${Q}^{''}$)\\
 }
 \Else
 {
 \textit{root} $\leftarrow \emptyset$\\
 \textbf{return} \textsc{Break}
 }
}
}
\textbf{return} \textit{root}
}
\caption{\textsc{ExpandTree}}
\label{alg: tree}
\end{algorithm}
\end{minipage}
\end{flushleft}

The \textsc{ExpandTree} method in Algorithm \ref{alg: tree} evaluates the task set ${Q}$ sequentially considering the order imposed by the depth to compute the tree expansion (line 2-3). For each $q \in {Q}$, the approach defines the initial state  ${I}$ using the node information (line 4) and modifies ${I}$ according to the outcome $u_{i}$ (line 5). ${P}_{TC}$ is updated with ${I}^{''}$ (line 6). Then, TraCE triggers the temporal planning search (line 7), which finds the largest plan solution considering the current initial state ${I}^{''}$. Here,  the branch is not the first; therefore $t_{\mathit{start}}$ is not 0. The \textsc{FindTemporalPlan} method finds sequences of actions that satisfy the propositional preconditions considering the (new) initial state information. Then, the planning approach uses  the TS method to schedule the action sequence considering  the starting time for the new plan is $t_{\mathit{start}} = n_{i}$.time. ${H}$ saves the right time information regarding each action in the temporal plan $\Pi_{T}$. Supposing  $\Pi_{T}$ exists (line 8),  the method creates a new branch $\textit{branch\_child}$ of the tree (new child) and finds the task queue ${Q}$ associated with the branch (line 9) using Algorithm \ref{alg: BuildBranch}. The branch is mapped (line 10) by labelling the outgoing edges from the node to its child, considering  $u_{i}$ information. The node children are then added to the tree (line 11). The strategy updates the task queue ${Q}$ to compute further branches considering the new sensing nodes identified in the $\textit{branch\_child}$ (line 12). If the approach does not find a plan  the \textsc{ExpandTree} returns an empty plan in $root$ (line 13-15).

We implement the tree expansion sequentially and all branches expanded must return a solvable temporal plan. The algorithm returns a $\textit{root}$ (line 16) when all branches are fully expanded. This occurs when the ${Q}$ set is empty and all possible branches are created. The $\textit{root}$ builds over the set of nodes explored during the tree generation. The node labelling and depth information are fundamental to obtaining the transition tree. Saving the times at which sensing actions are executed allows the strategy to maintain the action scheduling in all branches. The TraCE planner iterates the solution by considering possible updates in the initial deterministic plan solution. If the algorithm identifies an update in the $\Pi_{T}^{*}$ at the time the $\Pi_{TC}$ is achieved, the planner starts the process again. In this case, TraCE starts finding a new \textit{root} based on a new initial temporal plan solution. Notice that if an updated $\Pi_{T}^{*}$ exist, the process starts again independently on the plan in $root$. Therefore, if the last $root$ was empty, it does not affect the search for a new plan. For one planning iteration, if the computed plan presents a branching factor $\mu$ and its depth (maximum) is $\eta$, the tree describing $\Pi_{TC}$ has at most $\mu^{\eta}$ leaves, where the leaves are physical vertices. Therefore, the TraCE planner calls the temporal planning algorithm at most $\mu^{\eta}$ when finding a solution. TraCE provides a generic solution. The planner solves problems with temporal, numeric and perception requirements. However, it can also solve problems with fully-known initial states as a common temporal planning problem. Soundness and completeness results for  our approach are presented in Appendix \ref{appendix}.

\section{Experimental Evaluation}
Our domain and problems are encoded in PDDL.  All experiments in this section are run on Ubuntu 16.04, with an Intel Core i7-8700, limiting the planner to 30 minutes of CPU@3.2GHz,  16GB of RAM. We illustrate the TraCE planner performance in three experiments that cover offline planning and execution:

\noindent
\textbf{Experiment 1.} This experiment compares the planner with a state-of-art offline conditional planner PO-RPR, to evaluate the efficiency of the proposed approach generating solutions for a set of simple PPOS domains\footnote{The original problems and PO-PRP source are presented in \url{https://github.com/QuMuLab/planner-for-relevant-policies.git}}, including Coloured Balls (cballs), Canadian Traveller's Problem (ctp-ch), and Doors (doors). These domains are described in \cite{muise2014computing}.  TraCE uses a revised version of the domains considering durative actions; numeric constraints are not included.

\noindent
\textbf{Real-World Domains.} Experiment~2-3 considers planning domains motivated by real-world domains\footnote{These domains and problems can be found at \url{ https://github.com/YanielCarreno/tcp-domains}.}, particularly robotics applications, including the Offshore Energy Platform (oep), Manufacturing Plant (mp), Valve Manipulation (vm) and the Neighbourhoods (n) domains. All these domains describe robotics problems where temporal requirements (e.g., TILs and deadlines), numeric constraints, and partial observability are part of the mission characteristics.

\noindent
\textbf{Experiment 2.}  Here, we evaluate the quality of the planning algorithm while generating plans for real-world domains. The problems introduce a set of robotic scenarios where robots need to implement tasks with temporal and numeric constraints to maintain the operation, and sensing actions are required to solve the planing problem. 

\noindent
\textbf{Experiment 3.}  We evaluate the system's performance in a laboratory environment using a BlueROV2. The experiment analyses the mission execution time associated with plans that include non-deterministic sensing actions. Here the problem goals include the manipulation of five valves, sensors need to check the valves' state before executing the manipulation. We compare  TraCE  and OPTIC plan execution. The execution framework embedding both planners support replanning during the mission. The experiment evaluates ten different problems  where we introduce five forced failures during the mission, associated with the valve state and valve localisation.  

\begin{figure*}[!t]
\begin{minipage}{\textwidth}
\begin{minipage}[b]{0.38\textwidth}
\begin{center}
\scriptsize\addtolength{\tabcolsep}{-4pt}
\begin{tabular}{ccccc}
\toprule
\cmidrule{1-5}
\multicolumn{1}{c}{} & 
\multicolumn{2}{c}{\textbf{Size ($\Delta + {A}$)}}  & \multicolumn{2}{c}{\textbf{PT}} \\ \cmidrule{2-5}
%\midrule
%\toprule
\textbf{Problem}&
PO-PRP& TraCE &
PO-PRP& TraCE \\
\midrule
\rowcolor{almond}
\text{cballs-4-1}  &
 \text{261}&  \text{232} & 0.02 & \text{3.45}\\
\text{cballs-4-2}  &
\text{13887} & \text{TO} & \text{0.67} & \text{TO}   \\
\text{cballs-10-1}  &
 \text{4170} & \text{TO} & \text{1.57} & \text{TO} \\
 \rowcolor{almond}
\text{ctp-ch-1} &
\text{4} & \text{6} & \text{0.00} & \text{2.01}  \\
\rowcolor{almond}
\text{ctp-ch-5}  &
\text{16} & \text{16} & \text{0.00}& \text{7.05}  \\
\rowcolor{almond}
\text{ctp-ch-10} &
\text{31} & \text{35} & \text{0.02} & \text{7.10}   \\
\rowcolor{almond}
\text{ctp-ch-15}  & 
 \text{46} & \text{42}  & \text{0.07} & \text{8.24}  \\
 \rowcolor{almond}
\text{doors-5}  &
  \text{82} & \text{96} & \text{0.01} & \text{3.00} \\
\text{doors-7}  &
 \text{1295} & \text{2291} & \text{0.04} & \text{16.90}  \\
\text{doors-9}  &
 \text{28442} & \text{TO} & \text{1.07} & \text{TO}  \\
\bottomrule
\end{tabular}
\end{center}
\captionsetup{width=6.0cm}
\captionof{table}{Experiment 1: Plan size and planning time (PT) (sec)  for PO-PRP and TraCE when solving 10 simple PPOS problem instances. TO indicates time out.}
\label{tab: experiment1}
\end{minipage}
\begin{minipage}[b]{0.62\textwidth}
\begin{center}
\scriptsize\addtolength{\tabcolsep}{-4pt}
\begin{tabular}{ccccccccccccc}
\toprule
\cmidrule{1-13}
\multicolumn{1}{c}{} & 
\multicolumn{3}{c}{\textbf{oep}} & 
\multicolumn{3}{c}{\textbf{mp}}  & 
\multicolumn{3}{c}{\textbf{vm}}  & 
\multicolumn{3}{c}{\textbf{bc}} \\ 
\cmidrule{2-13}
\textbf{Problem}&
BF  & Size & PT & 
BF  & Size& PT & 
BF  &  Size& PT & 
BF  & Size&  PT \\
\midrule
\text{1}  &
\text{4}  & \text{85}  & \text{41.50}&
\text{2} & \text{78} & \text{32.30} &
\text{2} & \text{44} & \text{13.40} &
\text{2} & \text{42} & \text{9.05} \\
\text{2} &\text{3} & \text{77} &  \text{22.15}
& 3 & \text{57} & \text{47.50} &
\text{2}  & \text{64} &\text{28.12}  & 
 2 & \text{156} &  \text{28.63}\\
 \text{3} &
\text{2} &  \text{66} & \text{27.23}&
 2 &\text{108}   & \text{32.52}&
\text{2} & \text{67} & \text{21.23}& 
2 & \text{108} & \text{33.42}\\
\text{4} &
\text{2} &  \text{81}  &\text{32.03}&
4 &  \text{72} &\text{42.29} &
\text{2}  & \text{95} & \text{19.92} & 
 2 & \text{88}  &\text{14.81} \\
 \text{5} &
\text{2}  & \text{102} & \text{35.50} &
 3 & \text{340} &  \text{69.43}&
\text{2}  & \text{145} & \text{12.24} &
2 &  \text{234}& \text{44.70}\\
\text{6} &
\text{3}  &\text{90} & \text{39.34} &
4 &  \text{75}&  \text{52.18}&
\text{2}   & \text{132} &\text{14.14}  & 
 2 & \text{69}& \text{8.44}  \\
 \text{7} &
\text{3}  & \text{123}& \text{53.20} &
\text{3} & \text{72} &  \text{50.90} &
\text{2} & \text{80}& \text{17.15}  & 
 2 & \text{62} & \text{6.13}\\
 \text{8} &
\text{4}  & \text{116}& \text{66.23} &
\text{3} & \text{110} &  \text{84.32} &
\text{2} & \text{109}& \text{27.08}  & 
 2 & \text{92} & \text{19.13}\\
 \text{9} &
\text{5}  & \text{TO}& \text{TO} &
\text{4} & \text{145} &  \text{54.91} &
\text{2} & \text{56}& \text{18.83}  & 
 2 & \text{134} & \text{21.12}\\
 \text{10} &
\text{6}  & \text{TO}& \text{TO} &
\text{5} & \text{TO} &  \text{TO} &
\text{2} & \text{134}& \text{96.90}  & 
 2 & \text{287} & \text{48.32}\\
\bottomrule
\end{tabular}
\end{center}
\captionsetup{width=9.5cm}
\captionof{table}{Experiment 2: Branching factor (BF),  plan size (Size), and  planning time (sec) (PT), including physical and sensing actions, for 10 problem instances of the real-world robotic application domains using TraCE. TO indicates time out.}
\label{tab: experiment2}
\end{minipage}
  \end{minipage}
\end{figure*}

\section{Discussion}

The results for Experiment~1 are presented in Table \ref{tab: experiment1}. The TraCE planner finds a solution for most problem instances solved by PO-PRP in the experimental domains. The planner obtains plans with sizes similar to PO-PRP's plan sizes in several problems (highlighted in red). However, PO-PRP outperforms TraCE in terms of planning times. TraCE sequentially expands the tree, introducing a delay in plan generation. In addition, the TraCE planner uses temporal solvers to create the branches and expand the tree, which affects the planning times. Temporal planners must meet the causality and temporality constraints imposed by the model to obtain a valid plan. Therefore, additional time delays using TraCE compared to PO-PRP might be expected, as the PO-PRP planner does not consider the time to schedule the action sequence when finding a plan solution. We introduce time reasoning to address another set of problems that PO-PRP cannot solve. Here, we present the results for the first  $\Pi_{TC}$ obtained. Additional iterations optimise the plan solution by reducing the plan size; however, this introduces additional delays in the plan generation.
% We recognise that our planner's main strength is the capacity to solve complex problems with various constraint types (e.g., temporal, numeric, partial observability). For this experiment, PO-PRP addresses a single problem requirement (incomplete information), while TraCE finds solutions that additionally meet temporal constraints. Finally, we recognise 
%pledge the TraCE planner's main application area is robotics, where belief tracking is smaller than those in the benchmark contingent planning domains.

 Table \ref{tab: experiment2} shows that TraCE solves most of the problem instances. The planner generates solutions with different branching factors, demonstrating that TraCE deals with complex problems and scales well. As expected, we note that the size of the plans and the planning times rise with the increase in the number of sensing actions required in the plan to reduce the knowledge incompleteness.
 %to acquire the incomplete knowledge.
 The oep and mp domain incompleteness leads to more significant branching factors and planning times as the robots can explore multiple paths (for the oep domain) and control several flows in a valve (for the mp domain). TraCE planning times for the problem instances of the oep and mp domains are significantly higher compared to the other two domains. These domains require the robots to find plans that deal with concurrent action execution and numeric constraints along the branches which also affects planning times. The planner finds solutions for all problem instances of the vm and bc domains. Although those domains present small branching factors, planning times increase for complex problems (see problems 9 and 10). The main reason for this increase is that TraCE needs to find a solution for problems where the robot needs to recharge to maintain operation.
%The temporal solver requires then more time to accommodate all temporal constraints. The inclusion of reasoning regarding exogenous events (e.g., the time slot for recharging) that support the robot's recharge across different branches increases the planning time. The VM domain include concurrent action requirements (e.g., mapping and navigation). Implementing solutions for problems with concurrency requirements is an increase in the planning times. We can see this in problem instances 1-3, where TraCE consumes a significant amount of time finding solutions for problems with few goals. Finally,  Experiment~2 shows the robustness of the TraCE approach finding a plan that meets all temporal numeric and sensing constraints. 

\begin{figure}[!t]
\centering
\includegraphics[width=0.75\columnwidth]{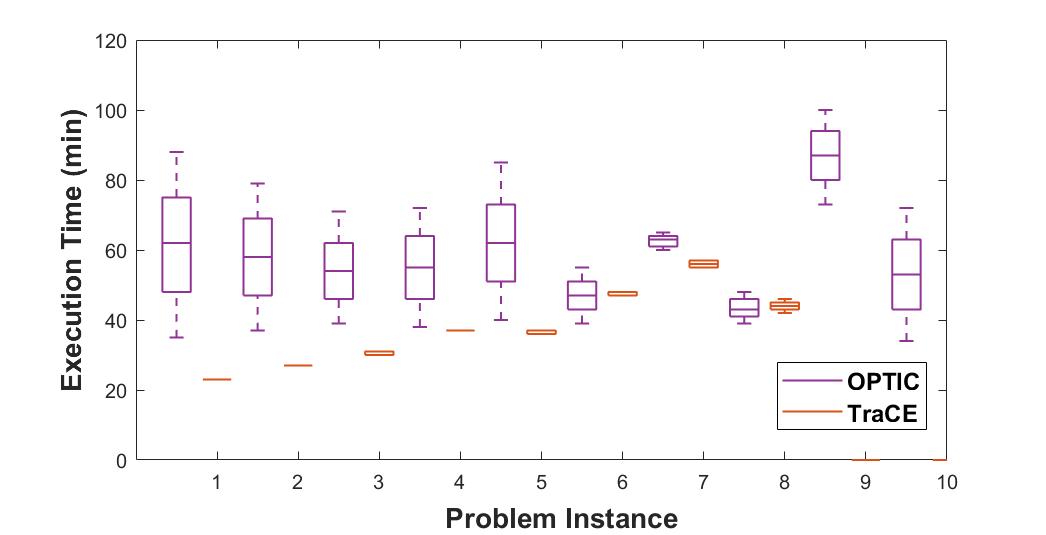}
\caption{Experiment 3: Plan execution time (min) in 10 problem instances of the  Example \ref{example} domain over 10 runs using TraCE and OPTIC. 
}
\label{fig: execution}
\end{figure}

Figure~\ref{fig: execution} shows that the generation of contingent branches during the planning stage significantly improves mission implementation times. This is mainly because OPTIC needs to replan every time a discrepancy between the real value from the sensor (run-time) and the expected outcome (planning time) is found. Replanning introduces unnecessary delays in the mission, mainly when we can use alternative algorithms that deal with certain levels of uncertainty in the domain. Our approach reduces replanning in non-quiescent environments since TraCE introduces the contingency analysis that specifies the possible outcomes from the observation.  $\Pi_{TC}$ can be affected by noisy sensors, which might force the system to generate a new plan if the sensing action does not provide a conclusive outcome.

\section{Conclusions}

We present the TraCE planner, which deals with incomplete knowledge, sensing, temporal notions, and numeric constraints. Our approach combines temporal plan construction in a contingent planning framework, offering more robust solutions for new types of applications. We can model problems that (i) require temporal reasoning, such as timed initial literals and deadlines;  (ii) manage resources, using numerical fluents, offering more powerful modelling of mission scenarios;  and (iii) and include partial observability.  We propose a PDDL extension to model  sensing actions that operate over unknown propositions, and  possible outcomes from non-deterministic sensing actions. We evaluate the approach in experiments that cover offline planning and execution in a new set of domains.

\nocite{*}
\bibliographystyle{eptcs}
\bibliography{references}

\appendix 
\newpage
\section{Soundness and Completeness}
\label{appendix}

%• Soundness
%– A planning algorithm is sound if all solutions are legal plans
%• All preconditions, goals, and any additional constraints are satisfied
%• Completeness
%– A planning algorithm is complete if a solution can be found
%whenever one actually exists
%– A planning algorithm is strictly complete if all solutions are
%included in the search space

%In theoretical computer science, an algorithm is correct with respect to a specification if it behaves as specified. Best explored is functional correctness, which refers to the input-output behavior of the algorithm

Temporal planning requires meeting the causality and scheduling constraints imposed by a temporal planning problem while looking for a sequence of actions that reaches the goal state. To achieve a valid plan, temporal planners must consider the action duration constraints along an unbounded length timeline. The TraCE planner includes the temporal planning algorithm \textsc{FindTemporalPlan}, which inherits the soundness and completeness of OPTIC and its predecessor POPF. The TraCE planner relies on constructing a transition tree where individual branches denote deterministic temporal plan solutions. This solution involves solving temporal planning problems with different initial states and merging them into a tree. Possible outcomes of nondeterministic sensing actions describe the initial states. Then, for a temporal plan solution:

\begin{definition}
\label{sound-plan}
A temporal plan is sound (with respect to a given domain and problem) if action preconditions are satisfied in their respective states, temporal constraints are met, and the final action produces a state where the goals are satisfied.
\end{definition}

\begin{definition}
\label{sound-temporal-plan}
A temporal planner (or temporal planning algorithm) is defined as sound if every temporal plan that the planner generates (with respect to a given domain and problem) is a sound plan.
\end{definition}

\begin{definition}
\label{complete-temporal-plan}
A planner is complete if for every planning domain/problem that has a solution, the planner is guaranteed to produce a plan.
%A temporal planner that computes a task plan whose plan size is at most ${Z}$ is ${Z}$-complete if it finds the task plan (if one exist)  or it finds out that there is no such plan. 
\end{definition}

%\begin{definition}
%\label{sound-temporal-plan}
%A temporal planner that can decide upon the actions needed to construct a valid plan considering a set of ordering constraints and the currently executing actions is a sound planner. 
%A temporal planner that can decide upon the actions needed to construct a plan considering a set of ordering constraints is a sound planner, where ordering constraints record that the action preconditions are satisfied and the ordered actions in the plan meet the set temporal constraints imposed by the problem. 
%\end{mydef}

%soundness I founded on the COLIN PAPER

The \textsc{FindTemporalPlan} temporal planning approach reasons about the state representation. The state considers: (i) the ordered list of start events (actions that have started but not yet finished) and the collection of temporal constraints over the actions in the plan to reach the current world state. These two state description components are defined over the sets ${P}$ and ${V}$ that hold in the world and support the planning approach's soundness.

We can explore the soundness and completeness of the TraCE planner considering it builds on the temporal plan solutions generated by the sound and complete method \textsc{FindTemporalPlan}. The temporal planning approach uses the domain actions to find a \emph{valid plan}---a plan obtained by a sound a complete planner%that meets all the problem constraints imposed by the domain definition making it a sound and complete plan
---that reaches the goal state. We assume the PDDL actions in the domain description do not present delayed effects. This means that action effects in the next state are defined in terms of the current state of the world.  Following these points, the TraCE planner soundness is defined as follows:

\begin{lemma}
(Soundness). TraCE is a sound planner if every branch of the tree from the root to a leaf constructed incrementally is a valid temporal plan of physical and sensing actions.
%(Soundness). Under the assumption the temporal planner is sound and complete, every branch from the root to a leaf incrementally constructed by TraCE is a valid temporal plan of durative physical and sensing actions.
%For a transition tree ${B}$   describing a $\Pi_{TC}$ every branch from the root to a leaf incrementally constructed by TraCE using a sound temporal planning strategy is a valid temporal plan of durative physical and sensing actions.
\end{lemma}

\medskip\noindent
\textbf{Proof.} Let $\Pi_{T}' = \langle a_{0}, a_{1}, \cdots, a_{n}\rangle$ be a temporal plan that solves the temporal planning problem ${P}_{T}$, where an action $a_{i}$ is a physical or a sensing action ($0 \leq i < n$) and the last action $a_{n}$ in  the plan is a physical action. This plan is valid considering Definition \ref{sound-temporal-plan} and Definition \ref{complete-temporal-plan}. Let ${H}' = \langle \langle{S}_{0}, t_{0}\rangle,  a_{0}, \langle{S}_{1}, t_{1}\rangle, $ $a_{1}, \cdots, \langle{S}_{n}, t_{n}\rangle, a_{n}, \langle{S}_{n+1},$ $t_{n+1}\rangle \rangle$ describe the history of $\Pi_{T}'$, where every action $a_{i}$ is executed at a belief state ${S}_{i}$ at a time $t_{i}$ and reaches a belief state ${S}_{i+1}$ at a time $t_{i+1}$ ($0 \leq i < n$). ${S}_{n+1}$ represents the last belief state that is the goal state considering  the problem ${P}_{T}$ definition.

For every sensing action $a_{j}$ in $\Pi_{T}'$ ($0 \leq j < n$) with an outcome $u_{j}$ observed  at $ \langle{S}_{j+1},t_{j+1}\rangle$, TraCE constructs a task $\langle n_{j}, u_{j} \rangle$ and introduces it into the queue of tasks ${Q}$. Let   $\Pi_{T}'' = \langle a_{j} = a_{0}', a_{1}', \cdots, a_{n'}'\rangle$ a temporal plan with an history ${H}'' = \langle \langle {S}_{j}, t_{j}\rangle = \langle{S}_{0}', t_{0}'\rangle,  a_{0}', \langle{S}_{1}', t_{1}'\rangle, a_{1}', \cdots, \langle{S}_{n'}', t_{n'}'\rangle, $ $ a_{n'}', \langle{S}_{n'+1}', t_{n'+1}'\rangle \rangle$ computed for the planning problem characterised by this task. This plan is valid considering Definition \ref{sound-temporal-plan} and Definition \ref{complete-temporal-plan}. Then, considering we are solving temporal planning problems with actions that do not present delayed effects, the sequence of actions $\langle a_{0}, a_{1}, \cdots, a_{j},a_{1}', \cdots, a_{n'}'\rangle$ is a valid sequential plan computed for a temporal planning problem ${P}_{T}$, with history $\langle \langle{S}_{0}, t_{0}\rangle, a_{0}, \langle{S}_{1}, t_{1}\rangle, a_{1}, \cdots, \langle{S}_{j}, t_{j}\rangle, a_{j}, \langle{S}_{1}', t_{1}'\rangle, $ $ a_{1}',
\cdots$ $,\langle{S}_{n'}', t_{n'}'\rangle, a_{n'}', \langle{S}_{n'+1}', t_{n'+1}'\rangle\rangle$. The first part of the plan, $\langle a_{0}, a_{1}, \cdots a_{j}\rangle$, does not prevent the last part of the plan, $\langle a_{1}', \cdots a_{n'}'\rangle$, and vice versa, considering $\Pi_{T}'$ and $\Pi_{T}''$ are sound and complete their union  is sound and complete.

% PLANNING PROBLEMS WITH DELAYED EFFECT CAN COME WITH THE INTRODUCTION OF TEMPORAL CONSTRAINTS!!!! 

We consider that all contingencies describing the incomplete knowledge in the initial state are specified in ${U}$. The TraCE planner uses this information to construct a tree with temporal branches sequentially.

\begin{lemma}
(Completeness). TraCE is a complete temporally-contingent planner if all known contingencies are specified by its input ${U}$, and incrementally constructs a temporal plan ${P}_{TC}$, such that the plan size of each sub-branch $b$ computed by one call of the temporal planning approach is smaller than the size of the first temporal plan obtained $\Pi^{*}_{T}$.
%(Completeness). \ac{TraCE} is a %${Z}$
%complete temporally-contingent planner if the  planner can compute all task plans, described by the sub-branches, that solve ${P}_{TC}$. The plan size of each sub-branch $b$ computed by one call of the temporal planning approach is smaller than the size of the first temporal plan obtained $\Pi^{*}_{T}$. 
\end{lemma}

\medskip\noindent
\textbf{Proof.} We first show that TraCE is complete considering (i) all known contingencies are specified by ${U}$,  and  (ii) the planner incrementally constructs a temporally contingent plan such that the size of each sub-branch $b$, computed by one call of the temporal planning approach is smaller than the size of the first temporal plan obtained $\Pi^{*}_{T}$. If all contingencies for a sensing action lead to a branch in the tree, then the tree cannot be extended further. Otherwise, for some sensing action contingency, there is no branch in the tree. This means that there does not exist a temporal plan whose plan's size is less than or equal to $\Pi^{*}_{T}$. Considering every temporal plan computed by \textsc{FindTemporalPlan} when building the tree ${B}$ is connected\footnote{Connection is guaranteed by ${H}$ and ${Q}$.}, such a branch cannot be reconstructed. If there is not a $\Pi_{TC}$ under the conditions we have mentioned in this proof, TraCE returns a failure. If there does not exist a temporally-contingent plan under the conditions and TraCE returns ${B}$,  at least one branch of the tree (excluding $\Pi^{*}_{T}$) from the root to a leaf is computed by one call of the temporally contingent planner. This leads to a contradiction when connecting the branches. Therefore TraCE either returns a complete temporally-contingent plan or failure.

In our approach, we analyse the plan size considering the number of actions in the plan. Other metrics such as the plan's makespan are not suitable for analysing the completeness in this case as we can find branches (described by a deterministic plan solution) with a small number of actions; however, with a higher makespan cost. For instance, consider $\Pi^{n}_{TC}$ denotes the solution for a $n$  robotic planning problem. The  $b_{1}$ and $b_{2}$ represent two branches in the plan with plan size 3 and 6, respectively. The branches' makespans are 40 min and 20 min, respectively. The first plan described by $b_{1}$ includes a navigation action with a significant duration. In this case, there is not a necessary correspondence between the two metrics. In our problems the makespan is ill-situated compare to the plan size in order to analyse the plan's branches. Following this reasoning, the plan size metric ensures we expand the tree from the most complex branches (based on the depth) to the simple ones, ensuring a solution where no further tree expansions are possible.

%\begin{lemma}
%(Correctness). If the solver generates plan ${{\Pi}_{TC}}$, the plan solves ${{P}_{TC}}$. If all tasks in ${Q}$ are expanded all branches were explored. The solution only exists if the approach successfully evaluates all nodes in the policy tree, where no nodes precedes its parents and children nodes associated with positive observations come first. 
%\end{lemma}

\medskip\noindent
\textbf{Plan Correctness.} When planning with temporal and contingent conditions, plan correctness relies on checking  (i) the temporal plans that build the branches are correct, (ii) the temporally-contingent plan satisfies all the goals, and (iii) the plan has sufficient knowledge at every point that supports its execution. The step (see Algorithm \ref{alg: TRACE}) in the TraCE algorithm \textit{CheckExistence} satisfies the first criterion considering  \textsc{FindTemporalPlan} returns a correct plan. The other two conditions are associated with checking the tree expansion (see Algorithm \ref{alg: tree}). If all tasks in ${Q}$ are expanded, all branches were explored. The solution only exists if the approach successfully evaluates all nodes in the  tree, where no nodes precede its parents and children nodes associated with positive observations come first. If ${Q}$ is empty the plan solution contains all the necessary knowledge to execute the sensing actions properly.

\end{document}